\documentclass{article}
\usepackage{arxiv}
\usepackage{graphicx}
\usepackage{hyperref}
\usepackage{url}
\usepackage{subcaption}
\usepackage{natbib}

\newcommand{\affit}[1]{$^{\mathrm{\textnormal{\textit{#1}}}}$}

\title{PDE foundation models are skillful AI weather emulators for the Martian atmosphere}

\author{
    Johannes Schmude\affit{a}
    \And Sujit Roy\affit{bc}
    \And Liping Wang\affit{d}
    \And Theodore van Kessel\affit{a}
    \And Levente Klein\affit{a}
    \And Marcus Freitag\affit{a}
    \And Eloisa Bentivegna\affit{a}
    \And Robert Manson-Sawko\affit{a}
    \And Bj\"orn L\"utjens\affit{a}
    \And Manil Maskey\affit{b}
    \And Campbell Watson\affit{a}
    \And Rahul Ramachandran\affit{b}
    \And Juan Bernabe-Moreno\affit{a}
}

\begin{document}

\maketitle

\begin{abstract}
We show that AI foundation models that are pretrained on numerical solutions to a diverse corpus of partial differential equations can be adapted and fine-tuned to obtain skillful predictive weather emulators for the Martian atmosphere. We base our work on the Poseidon PDE foundation model for two-dimensional systems. We develop a method to extend Poseidon from two to three dimensions while keeping the pretraining information. Moreover, we investigate the performance of the model in the presence of sparse initial conditions. Our results make use of four Martian years (approx.~34 GB) of training data and a median compute budget of 13 GPU hours. We find that the combination of pretraining and model extension yields a performance increase of 34.4\% on a held-out year. This shows that PDEs-FMs can not only approximate solutions to (other) PDEs but also anchor models for real-world problems with complex interactions that lack a sufficient amount of training data or a suitable compute budget.
\end{abstract}

\footnotetext[1]{IBM Research}
\footnotetext[2]{NASA Marshall Space Flight Center, Huntsville, Al, USA}
\footnotetext[3]{Earth System Science Center, University of Alabama in Huntsville, Al, USA}
\footnotetext[4]{Department of Electrical and Computer Engineering, Colorado State University, Fort Collins, CO, USA}

\section{Introduction}

Recent years have seen massive progress in the performance and adoption of AI emulators for atmospheric physics. Starting from the work of \citet{pathak2022fourcastnet} papers such as \citep{keisler2022forecasting}, \citep{bi2023accurate}, \citep{lam2023learning} have shown that purely data-driven AI models achieve a performance that rivals that of operational numerical weather prediction (NWP) systems running on HPC infrastructure rather than GPU. Typically the authors report a large increase in compute efficiency compared to the conventional NWP/HPC approach. By now, many national and international meteorological organizations have started developing their own models \citep{lang2024aifs,abdi2025hrrrcast}. While the above papers strictly focused on the forecasting problem, this development did heavily overlap with the foundation model revolution in language \citep{bommasani2021opportunities} and beyond. Thus it should be not surprising that a number of weather foundation models arose that aimed to demonstrate generalizability across a variety of tasks. Thse can include a series of challenging forecasts \citep{bodnar2025foundation}, but also problems such as downscaling or parametrizations \citep{lessig2023atmorep, schmude2024prithvi}.

In either case, what all these papers have in common is that they rely on a long time series of historic weather data to train on. Typically this is a global reanalysis such as ERA5 \citep{hersbach2020era5} or MERRA-2 \citep{gelaro2017modern}. One of the questions we are interested in is what happens if that data is not available. Mars atmospheric dynamics, for example, pose a data-limited problem due to the challenges in retrieving long-term historical measurements and associated reanalyses.

A simple and direct way out of this conundrum is to generate more data. However, this seems backwards. One sets up an HPC model to generate data to train and AI model so that one does not need to run a HPC model. An alternative is to embed more information in the model architecture. That way, only parts of the dynamic is learned. Although not framed in terms of data efficiencies, this is what happens in \citet{kochkov2024neural}. Here, the authors embed a hard-coded yet autograd-enabled dynamic core in an AI model, allowing to combine known with learned physics. Another alternative would be a transfer learning approach. Across scales \citep{nipen2025regional,nordhagen2025high} or regions. We ask the question whether rather than transfering information from one meteorological system to another, one can use knowledge of a variety of complex \emph{dynamical systems} as the starting point and anchor.

In particular, we are interested in the question whether so-called PDE Foundation Models (PDE-FMs) are suitable to counter the data appetite of data-driven models for atmospheric phyiscs. The general concept of PDE-FMs is the follows: One trains a model, typically some sort of vision transformer, on a large set of numerical solutions of a curated set of partial differential equations. The aim is to show data-efficient application to different PDEs. Naturally, the more those different PDEs deviate from what was seen at training time -- equation type, parameters, initial and boundary conditions -- the better. In a way this is a fusion of work on neural operators with the FM paradigm. Recent examples are \citep{herde2024poseidon,rautela2025morph,mccabe2025walrus}.

\paragraph{Contributions}
To our knowledge, this is the first study of applying the PDE foundation model approach to a complex real-world problem such as atmospheric physics. In this setting, our results show that pretraining on a diverse corpus of solutions to PDEs unlocks data and compute efficiencies for atmospheric physics models. Moreover, we introduce a technique to extend models pretrained on 2D data into a third spatial dimension. We also show that the above observations hold in the presence of sparse input data. Our results also point to ways to improve PDE-FMs to improve transition to and application in the atmospheric physics setting. We will discuss so in detail in section \ref{sec:discussion}.

\section{Background}

\subsection{AI models for atmospheric physics}
\label{sec:background:ai_for_atmospheric_physics}

When applying purely data-driven methods to atmospheric physics and medium range weather prediction, the work of \citet{pathak2022fourcastnet}, \citet{keisler2022forecasting}, \citet{bi2023accurate} and \citet{lam2023learning} and others has led to what is essentially a standard, baseline recipe. One starts with a long time series of global reanalysis data such as ERA5 \citep{hersbach2020era5} or MERRA-2 \citep{gelaro2017modern}. From these datasets one selects about four parameters at surface and six at vertical levels.\footnote{Typically, surface level parameters are 2 meter surface temperature, 10 meter u and v wind components as well as mean sea level pressure. At vertical levels one uses temperature, geopotential and humidity as well as three wind components.} Furthermore, one subsets the vertical dimension to a bit more than a dozen levels. Finally, there are auxiliary inputs such as orography, land-sea masks, top of atmosphere radiation, latitude, longitude, time of day and day of year. Using two timestamps of data as well as the static information, one then regresses on the same (non-static) variables six hours in the future. Models typically work with a fixed time step and reach longer lead times by repeat application of the model; a process referred to as \emph{autoregressive rollout}. Improving this recipe is a very active and fruitful field of research. Some papers extend the above to the regional setting, while others are concerned with obtaining stability at longer and longer forecast horizons. Others focus on extremes or difficult to predict variables such as precipitation.

For the purpose of our discussion, it is especially important to note that the above approach is somewhat data hungry. The exact amount of data required to train a model depends on the domain. While global models typically use 40 years of reanalysis data, regional ones use as little as three years \citep{andrychowicz2023deep,abdi2025hrrrcast,nipen2025regional}. Whatever the exact requirement, there are many cases where such data is not available. Not every part on earth is covered by a high resolution (re-) analysis project that is archived and available for deep learning. And as we will see in section \ref{sec:methods:data}, the OpenMARS database \citep{holmes2020openmars} comprises about eight Martian years.

As written in the introduction, one of the aims of this study is to understand what other information -- in particular data -- can be used to train a data-driven model. While the problem exists both in the regional setting on Earth as well as the global setting on Mars, we focus on the latter as the dynamics are arguably simpler: There is no humidity in the fundamental equations and no atmosphere-ocean interaction.

\subsection{PDE FMs}
\label{sec:background:pde_fms}
Under the foundation model paradigm \citep{bommasani2021opportunities} one aims to \emph{pretrain} a model in a generic, task-independent manner and subsequently fine-tune it to a series of downstream tasks. The point is to unlock data or compute efficiencies or achieve superior accuracies. The biggest successes have unarguably been in the areas of language and vision. However, this paradigm has been applied to many other domains, including weather \citep{lessig2023atmorep,nguyen2023climax,schmude2024prithvi,bodnar2025foundation}.

\citet{herde2024poseidon} takes this to the setting of partial differential equations. The model, \emph{Poseidon}, is pretrained on numeric solutions to a set of PDEs. And the authors find that it can be effectively fine-tuned to solve previously unseen PDEs. See also \citep{rautela2025morph,mccabe2025walrus}. Note that from an AI perspective, these models are very, very similar to the weather models discussed in section \ref{sec:background:ai_for_atmospheric_physics}. That is one takes the state of the system at one or several timestamps as input and regresses onto a future timestamp. After initial one-step-ahead training, the lead time is improved via autoregressive rollout.

\subsection{Poseidon}
\label{sec:background:poseidon}

As we will be leveraging Poseidon for our experiments, let us go over the most relevant aspects of its architecture. The model underlying Poseidon is the \emph{scalable operator Transformer} (scOT). scOT is essentially a Swin V2 U-Net \citep{liu2022swin} with ConvNeXt layers \citep{liu2022convnet} in its residual connections. All inputs and outputs have four channels representing density $\rho$, horizontal and vertical velocity ($u$, $v$) as well as pressure $p$. If a variable is not part of a given PDE, it is set to zero. In addition, the model is strictly two-dimensional and was pretrained with data measuring $128 \times 128$ pixel.\footnote{We use Poseidon-B, which has a patch size of $4 \times 4$ and four U-Net stages, meaning that the pixel count along each axis has to be a multiple of 64.} In contrast to most weather models which leverage two timestamps as input, Poseidon only uses a single timestamp as initial condition. It was pretrained with numeric solutions to the Navier-Stokes and Compressible Euler equations. While several of the downstream tasks in \citet{herde2024poseidon} use non-periodic boundary conditions, all pretraining data is subject to periodic ones. Indeed, one of the findings of \citet{herde2024poseidon} is that the model generalizes well to different boundary conditions at fine-tuning time.

\section{Methods}

\subsection{The OpenMARS dataset}
\label{sec:methods:data}

We run all our experiments with reanalysis data from version 5 of the OpenMARS database \citep{holmes2020openmars}. In its original form, the data has a resolution of five degrees resulting in $36 \times 72$ pixels on a regular lat/lon grid. There are 35 vertical sigma levels. The vertical coordinate is defined in terms of pressure relative to surface pressure $p_s$, $\sigma = p / p_s$. The highest level in the dataset, $\sigma=5.0824954 \times 10^{-5}$, lies at an altitude of about $105$ km. The dataset contains temperature ($T$) as well as easterly ($u$) and northerly ($v$) wind as vertical variables. At surface level we have surface pressure, surface temperature and surface CO2 ice. Finally, there is the optical depth through the atmosphere resulting from dust. For our experiments, we only leverage $u$, $v$, $T$ at sigma levels. Moreover, we train form sol $2674.416748$ to sol $5348.750000$ and validate using sol $5348.833496$ to $6031.000000$. This corresponds to Mars Years 28 through 31 for training and Mars Year 32 for validation. In principle the OpenMARS dataset contains data from Mars Years 28 through 35.

To show the impact of pre-training, we run initial experiments on a single sigma level. Subsequent experiments across the vertical column. In the latter case we use essentially every second sigma level. This amounts to the configuration shown in table \ref{tab:vertical_levels}.

\begin{table}[htbp]
\caption{Sigma level choices for different experiments}
\label{tab:vertical_levels}
\begin{center}
\begin{tabular}{ll}
\multicolumn{1}{c}{\bf EXPERIMENT}  &\multicolumn{1}{c}{\bf LEVELS}
\\ \hline \\
Single level     &0.87007517. \\
Multi level   &0.9995, 0.99636, 0.9892728, 0.97349006, 0.93936884, \\
           &0.87007517, 0.74549896, 0.5632723, 0.36277717, 0.20065443, \\
           &0.099399455, 0.04613231, 0.020647187, 0.009001522, 0.0037814784, \\
           &0.0014586191, 0.00044534708, 5.0824954e-05.
\end{tabular}
\end{center}
\end{table}

\subsection{Data processing}
\label{sec:methods:data-process}

To match the fixed resolution and aspect ratio of the Poseidon datasets, we interpolate the OpenMars data to the same resolution -- $128 \times 128$ pixel -- using spline interpolation. Given that the original order of channels in the pretraining data was $\rho$, $u$, $v$, $p$, we order the OpenMars channels as $T$, $u$, $v$; inputs for $p$ are generally set to zero. We apply conventional standard scaling. Scaling parameters depend on channel and level; yet not on latitude or longitude. The statistics are computed from the training data.

\subsection{Training and fine-tuning}
\label{sec:methods:training-and-tuning}

When training or fine-tuning, we use batch sizes of 15 for single level and 10 for multi sigma-level data. The difference is motivated by GPU memory utilization. Except where noted, we train or tune for 20,000 gradient descent steps on a single A100 or H100 GPU. Exact training time varies depending on cluster and node utilization, yet our experiments for training a model with 18 vertical levels run at a median time of 13 GPU hours. We observed a minimal training time of 11.8 GPU hours.\footnote{These numbers represent wall time and include validation and testing loops, start-up, checkpointing, etc. If one truly wants to minimize GPU utilization, one can probably push for further efficiencies.} We always use the Poseidon-B configuration of the model which comprises about 158 million parameters. We train and tune using the same loss function as in \citet{herde2024poseidon}, a normalized L1-loss:
\begin{equation}
    L(\hat{x}, x) = \frac{1}{C} \sum_c \frac{\sum_{b, h, w} \vert \hat{x}_{bchw} - x_{bchw} \vert}{\sum_{b, h, w} \vert x_{bchw} \vert + \epsilon}.
\end{equation}
Here, $c$ sums over channels $C$ while $b$, $h$ and $w$ index the batch, latitude and longitude respectively. Finally, $\epsilon=10^{-10}$. As in \citet{herde2024poseidon}, we use cosine annealing and an initial learning rate of $10^{-5}$.

For the 3D models with vertical dependence we run inference across all 18 levels, yet randomly sample 9 adjacent levels from table \ref{tab:vertical_levels} at each training step. This alleviates memory pressure that would result from running on the entire 18 levels vertical.

As mentioned in section \ref{sec:background:poseidon}, the pretraining data for Poseidon uses periodic boundary conditions along both axes. Apart from the fact that \citet{herde2024poseidon} report strong performance when fine-tuning to different boundary conditions, one should also consider the following: Poseidon-B uses a patch size of four. Data inputs measure $128 \times 128$ pixel or $32 \times 32$ tokens. Given a window size of sixteen, every single Swin stage apart from the outermost ones computes effectively global attention. Moreover, the ConvNeXt blocks use zero padding.

\subsection{Extension to three dimensions}
\label{sec:methods:3d-extension}

\begin{figure}[htb]
\begin{center}
\includegraphics[width=1.0\textwidth]{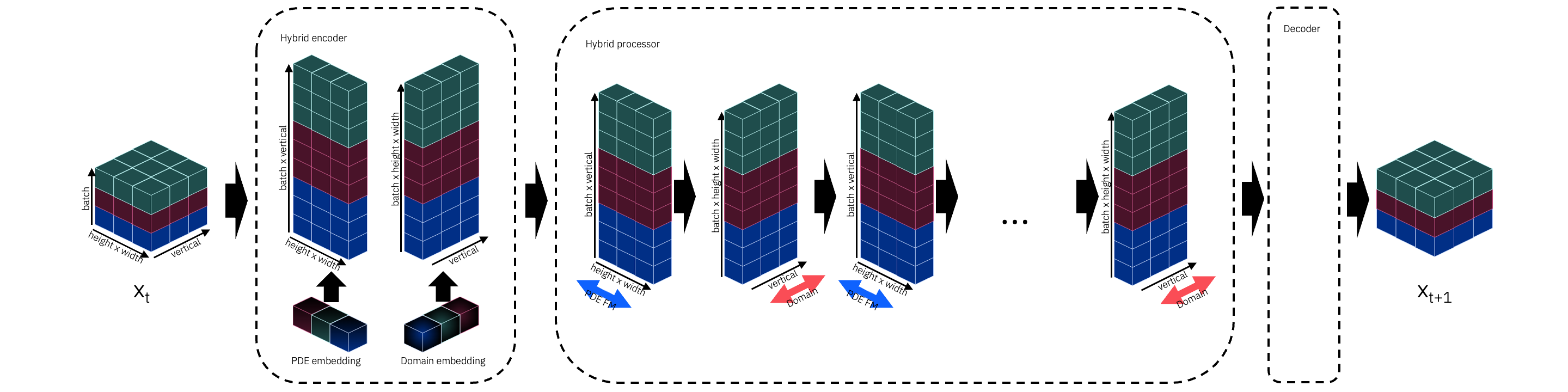}
\end{center}
\caption{Extending scOT to three dimension}
\end{figure}

We adapt the scOT architecture as follows to leverage the model pretrained on 2D PDE data for the 3D atmosphere of Mars: If we denote the shape of the 3D data as $(B, D, H, W, C)$, with $B$ the batch, $C$ the channel dimension and $D$, $H$, $W$ being depth, height and width, we reshape the data as $(B \times D, H, W, C)$ for any pretrained layer from the original scOT architecture. That is, we hide the vertical dimension in the batch dimension. While this allows us to use scOT on 3D data, it does not allow for any information flow along the vertical. Thus, we introduce additional transformer layers. These layers act along $D$ alone while $H$ and $W$ are now being rolled into the batch dimension. Essentially, this amounts to axial attention in a generalized 2D space: One ``axis'' given by $H \times W$ and the other by $D$. The layers for vertical attention are randomly initialized. This is quite similar to the temporal extension of \citet{blattmann2023align} or \citet{guo2024animatediff}. This new vertical attention acts at the end of every Swin layer; we do not augment the ConvNeXt blocks in any way. For Poseidon-B, this adds about 100M parameters to the model. That is, the combined scOT model with attention along the vertical comprises about 258M parameters; 158M stem from Poseidon-B. When fine-tuning, the 100M new parameters are randomly initialized. As outlined above, we typically train on only nine adjacent vertical levels while running inference across 18. Naturally, this is a byproduct of using attention to handle the vertical interaction rather than something else.

The axial attention along the vertical requires a suitable additional position encoding. Instead of a level-specific learned embedding or a Fourier-type one, we learn a function that maps the sigma coordinate to a suitable embedding. This function is a two-layer MLP with GELU activation. This means that we can run inference on levels not seen at training time.

\subsection{Sparsity}
\label{sec:methods:sparsity}

In certain experiments we will sparsify the data. This means sampling a random set of lat/lon locations for which we will either keep or discard the entire vertical column. When doing so, we will no longer set the fourth channel (originally trained as pressure $p$) to zero yet instead fill it with a binary mask indicating absence or presence of data.

\subsection{Baselines}
\label{sec:methods:baselines}

We will consider two baselines: Persistence and scOT initialized from random. Apart from the additional ConvNeXt layers and the fact that it uses a single timestamp as input, the scOT architecture is not atypical to what one finds in a number of weather models: Swin transformer layers arranged as a UNet. Thus we argue that scOT is quite representative of a generic AI model for weather. And we use the baseline twofold: To establish the efficacy of fine-tuning; yet also to compare against a weather model.

\section{Experiments}

\begin{table}[htbp]
\caption{Efficacy of PDE Foundation Models for atmospheric physics tasks. Loss in model units. The ``mixed'' initialization uses the PDE-FM weights on the 2D layers while randomly initializing the vertically acting layers.}
\label{tab:results:dense}
\begin{center}
\begin{tabular}{lrr}
\multicolumn{1}{c}{\bf MODEL INIT.}  &\multicolumn{1}{c}{\bf SINGLE LEVEL} &\multicolumn{1}{c}{\bf FULL VERTICAL}
\\ \hline \\
Random         &$0.078625$ &$0.142879$ \\
PDE-FM (ours)        &$0.062886$ &N/A\\
Mixed (ours)          &N/A      &$0.093776$\\
Improvement &$20.0\%$ & $34.4\%$
\end{tabular}
\end{center}
\end{table}

\subsection{Efficacy of multi-PDE pretraining}

\subsubsection{Single sigma level}
\label{sec:experiments:single-level}

To test the applicability of the PDE FM approach to the setting of atmospheric physics on Mars, we initially test training on a single sigma level. The rationale is simple: This is the closest to the pretraining setting of \citet{herde2024poseidon}. Indeed, Poseidon was trained on 2D data, so we restrict to a 2D setting. The results of this -- in terms of validation losses -- can be seen in the column ``single level'' in table \ref{tab:results:dense}. I.e.~there is a $30.1\%$ improvement. There is one important caveat here: Such improvement numbers are meaningless without a statement of physical quantities and performance. I.e.~if the model error for temperature would be $200K$, the improvement is of limited value. We will therefore later give errors for physical quantities as well.

\subsubsection{Full vertical}
\label{sec:experiments:full-vertical}

Before doing so, we discuss the performance when extending the model along the vertical however. Once again, results can be found in table \ref{tab:results:dense}. Recall that the model now has additional attention blocks and embedding layers acting along the vertical. We find an improvement of $34.4\%$ by pretraining. Note that the validation scores in table \ref{tab:results:dense} are averaged across levels. Thus, the scores for ``single level'' are computed at $0.87007517$, while those for ``full vertical'' are computed across all 18 levels of table \ref{tab:vertical_levels}. Thus, the numbers should not be directly compared.

\begin{figure}[htbp]
\begin{center}
\includegraphics[width=0.5\textwidth]{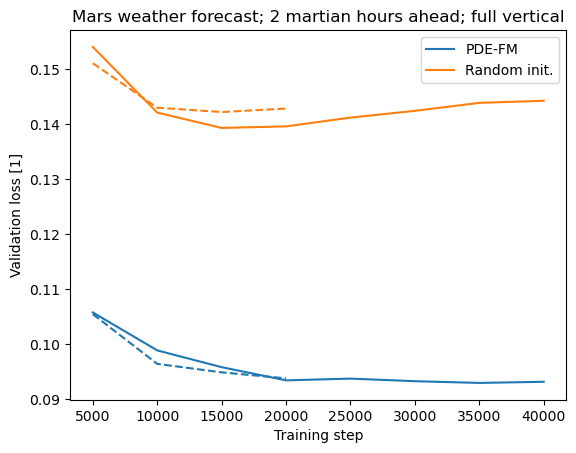}
\end{center}
\caption{When extending the training to 40,000 steps, the randomly initialized model not only does not close the performance gap, but rather shows clear signs of overfitting. The differences between the 20,000 and 40,000 step runs stem from the learning rate scheduler.}
\label{fig:overfitting}
\end{figure}

At this stage the reader might ask how these improvements change if we increase the available computational resources. Indeed, in the limit of infinite compute and infinite data the impact of pretraining data -- whether from other atmospheric systems or from general PDE data -- has to be zero. Yet we are particularly interested in the case of limited data or compute resources. Thus, we ran the 3D configuration of the model (i.e.~the one with additional attention blocks acting across up to 18 sigma levels) out to 40,000 gradient descent steps. The validation curves across this can be seen in figure \ref{fig:overfitting}. As should be clear from the figure, the randomly initialized model not only fails to close the performance gap; it actually shows clear signs of overfitting while the pretrained configuration remains stable.

\subsubsection{Sparsification}

As a final experiment, we study the impact of removing data. Following our discussion in section \ref{sec:methods:sparsity}, we do so by randomly dropping entire columns and indicating absence/presence of data with a binary mask in the $p$ channel of Poseidon. The loss is computed across all pixels, whether dropped or not. This follows the methodology of \citet{schmude2024prithvi}, the rationale being that all pixels -- masked or unmasked -- change as we are predicting the state into the future. As an implementation note, we drop columns randomly for each sample. That is, if the model sees a sample twice during training, it will do so with a different masking patterns. This is different to a scenario where one wanted to optimize for some hypothetical weather stations approximated via a sparsity pattern. In this case, the masking would be constant or follow some other predetermined pattern. We sample uniformly across latitudes and longitudes. In either case, results in terms of validation loss are listed in table \ref{tab:results:sparse}. As one can see, the pretraining once again significantly boosts model performance.

\begin{table}[htbp]
\caption{Efficacy of multi PDE pre-training in presence of missing data.}
\label{tab:results:sparse}
\begin{center}
\begin{tabular}{lrr}
\multicolumn{1}{c}{\bf MODEL INIT.}  &\multicolumn{1}{c}{\bf 80\%} &\multicolumn{1}{c}{\bf 95\%}
\\ \hline \\
Random &$0.161899$ &$0.179392$ \\
PDE-FM (ours) &$0.098258$ &$0.118802$\\
Improvement &$39.3\%$ &$33.8\%$
\end{tabular}
\end{center}
\end{table}

\subsection{Physical variables}

\begin{figure}[p]
\begin{center}
\includegraphics[width=0.6\textwidth]{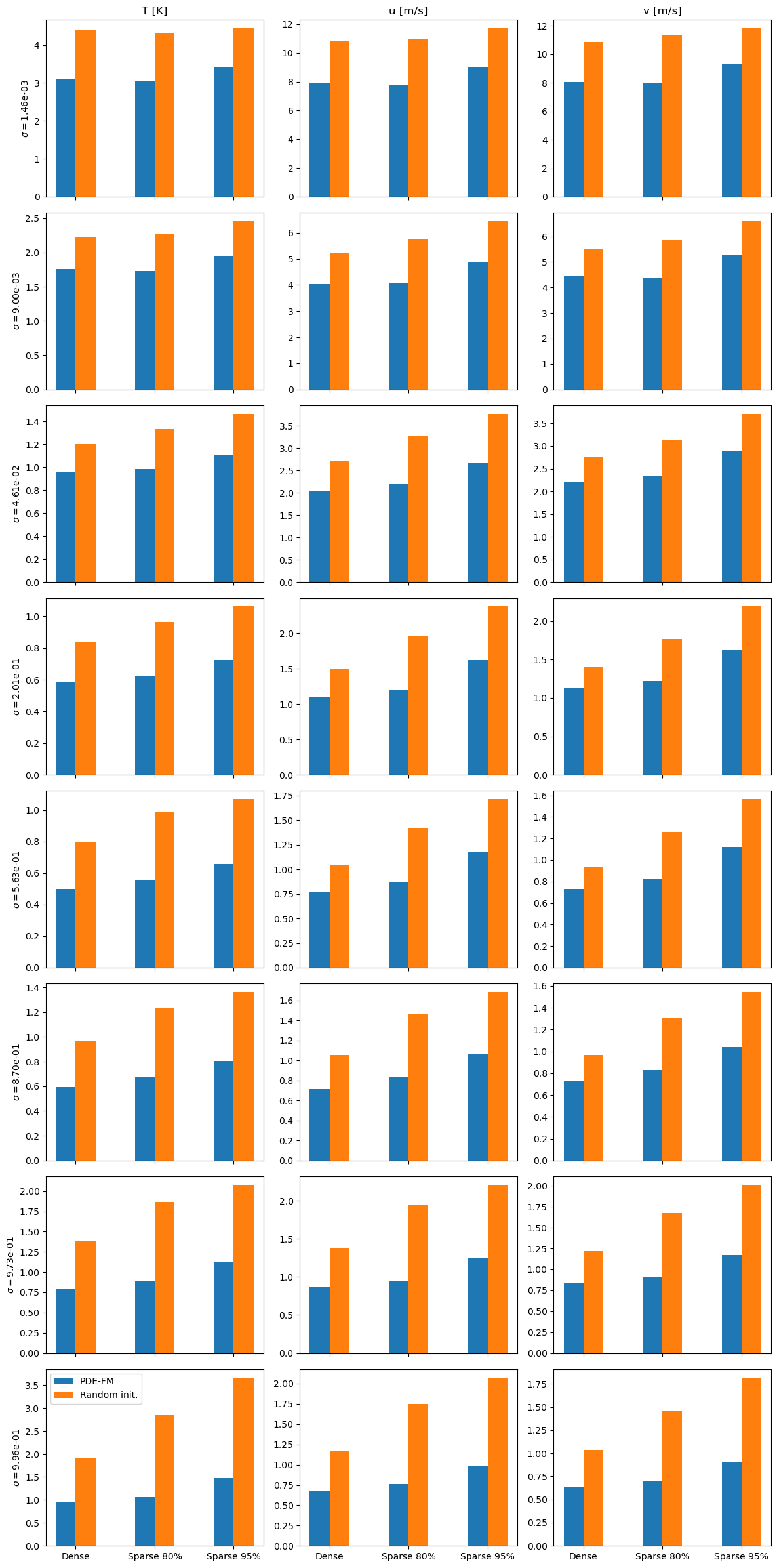}  
\end{center}
\caption{Model performance after 20,000 training steps by level and masking ratio.}
\label{fig:sparse_dense_results}
\end{figure}

\begin{figure}[htbp]
\begin{center}
\begin{subfigure}{.4\textwidth}
        \centering
        \includegraphics[width=0.8\linewidth]{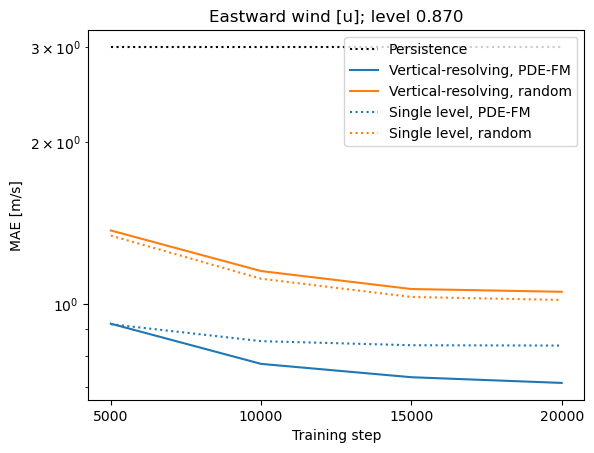}  
        \caption{Eastward wind}
        \label{fig:mars_3d_forecast_impact_eastward_wind}
    \end{subfigure}
    \begin{subfigure}{.4\textwidth}
        \centering
        \includegraphics[width=0.8\linewidth]{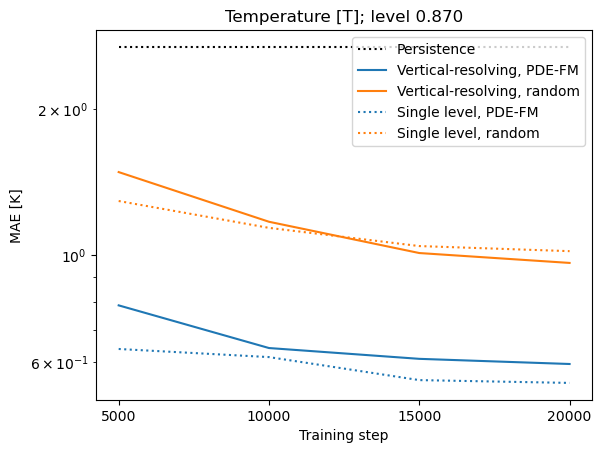}        
        \caption{Temperature}
        \label{fig:mars_3d_forecast_impact_temperature}
    \end{subfigure}
\end{center}
\caption{Impact of pretraining and dimensional extension on physical quantities.}
\end{figure}

As alluded to in section \ref{sec:experiments:single-level}, the acccuracy scores provided so far need to be accompanied with quantities that can be physically interpreted. For a clear comparison across our different models and training scenarios, we show MSEs at the $0.87007517$ sigma level for pre-trained and fine-tuned models and the 2D and 3D configurations as well as persistence in figures \ref{fig:mars_3d_forecast_impact_eastward_wind} and \ref{fig:mars_3d_forecast_impact_temperature}. Several points should be quite clear from the plots: To start, all models outperform the persistence baseline significantly. The 3D models improve the performance compared to their 2D counterparts regarding wind speed while sacrificing accuracy for temperature -- at least at this level. One gains about $14.8\%$ for eastward and $14.1$ for northward wind speed while sacrificing $9.4\%$ for temperature at this level. Finally, while there are few baselines for AI weather models for Mars, one can make a rough comparison by looking at Earth. The obvious way to do so is via the Weatherbench 2 benchmark \citep{rasp2024weatherbench}. There, for a resolution of 32 by 64 pixel one finds for temperature at $850 hPa$ at six hours ahead a persistence error of $1.16K$; the IFS HRES model is listed as $0.24K$. This is an improvement of about $80\%$ which is roughly what we are seeing as well. This is not to argue that this is peak performance for atmospheric physics on Mars; but to show that the accuracy gains from pretraining and dimensionality adaption in sections \ref{sec:experiments:single-level} and \ref{sec:experiments:full-vertical} lead to scores which are of the correct order of magnitude to be competitive.

Let us now turn to the vertical as well as the question of sparsification. Figure \ref{fig:sparse_dense_results} shows MAEs for temperatur and both wind speeds at eight different levels for masking ratios of $0\%$, $80\%$ and $95\%$. Regarding the vertical, we see lowest MAEs at the lower yet not lowest levels. While we do not know the \emph{exact} reason for this behavior, it is important to point out that our model has no knowledge of orography, nor of incoming radiation. Consider the \emph{standard recipe for AI weather emulators} outlined in section \ref{sec:background:ai_for_atmospheric_physics}. There, we mentioned that typical models not only include orography and top of atmosphere radiation; yet also information on latitude, longitude as well as time from which they can infer radiative parameters and regional characteristics. With this in mind, it would stand to reason that our model would perform poorly at the highest and lowest level; given that it has no way to infer the correct boundary conditions. Having said that, model error is also expected to increase with higher levels. So one should take the above discussion with a grain of salt.

Another important feature of figure \ref{fig:sparse_dense_results} is that the loss of accuracy with diminishing data density is \emph{considerably} less pronounced in the pretrained model.

\begin{figure}[htb]
\begin{center}
\includegraphics[width=0.8\textwidth]{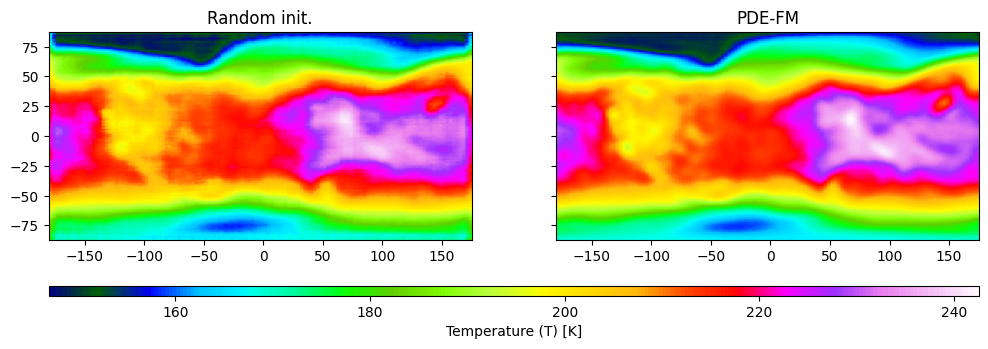}  
\end{center}
\caption{Sample prediction for temperature at $\sigma=0.87007517$.}
\label{fig:forecast_T}
\end{figure}

\begin{figure}[htb]
\begin{center}
\includegraphics[width=0.8\textwidth]{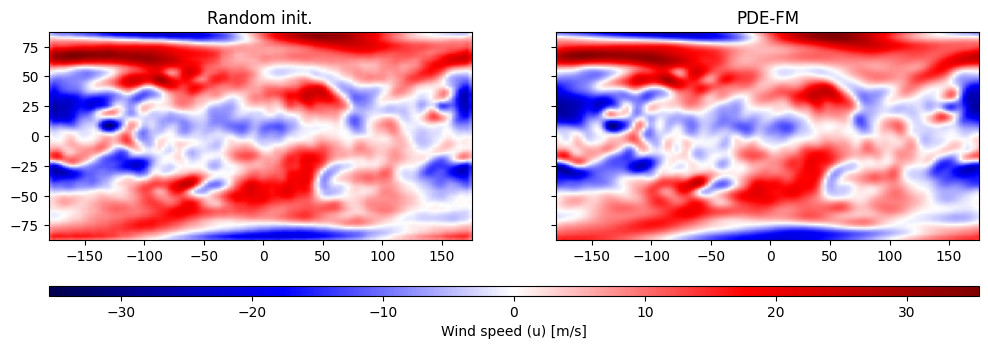}  
\end{center}
\caption{Sample prediction for u component of wind at $\sigma=0.87007517$.}
\label{fig:forecast_u}
\end{figure}

\begin{figure}[htb]
\begin{center}
\includegraphics[width=0.8\textwidth]{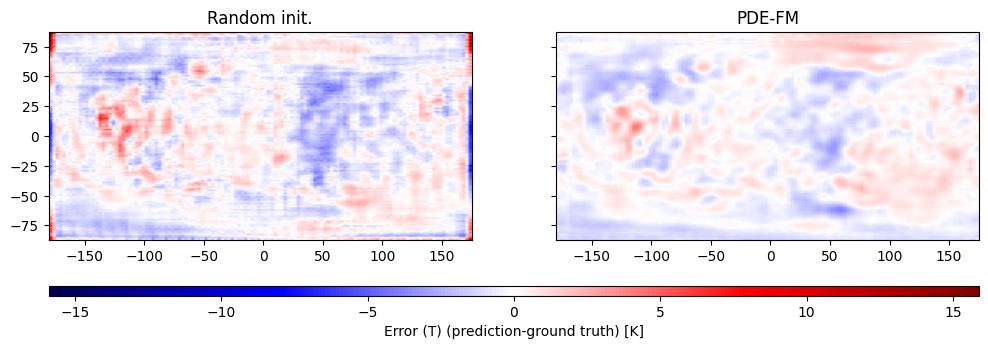}  
\end{center}
\caption{Forecast error for temperature at $\sigma=0.87007517$.}
\label{fig:forecast_error_T}
\end{figure}

\begin{figure}[htb]
\begin{center}
\includegraphics[width=0.8\textwidth]{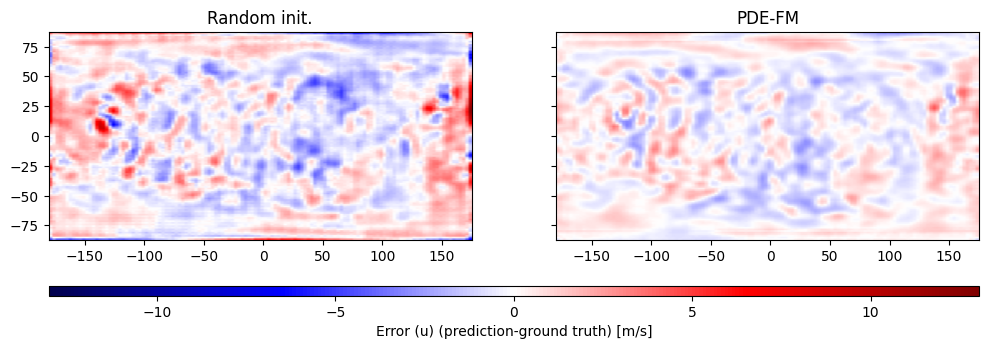}  
\end{center}
\caption{Forecast error for u component of wind at $\sigma=0.87007517$.}
\label{fig:forecast_error_u}
\end{figure}

We conclude our discussion by considering the spatial patterns in forecasts as well as forecast errors. The former are plotted in figures \ref{fig:forecast_T} and \ref{fig:forecast_u}; the latter in figures \ref{fig:forecast_error_T} and \ref{fig:forecast_error_u}. Both errors show a clear boundary effect in the randomly initialized model. Given the architecture of scOT, this is most likely due to the zero padding in the depth-wise convolutions in the ConvNeXt layers. Beyond this, especially the temperature error of the randomly initialized model also exhibits what are probably tokenization artifacts. Once again, the interpretation is up for debate, but it would stand to reason that the pretraining has introduced a strong bias for local dynamics into the transformer architecture.

\section{Discussion}
\label{sec:discussion}

The main question of this study was whether PDE foundation models are a suitable approach to deal with data poor settings in atmospheric physics. Our results, in particular figure \ref{fig:overfitting}, answer this question to the positive. At least for the data and architecture considered here, it would have not been possible to obatin comparable results by training from scratch as overfitting sets in way before that. Naturally, one can wonder whether regularization using a much leaner architecture or conventional regularizers such as dropout etc.~would lead to a different conclusion.

It is important to interpret this finding with the architectural characteristics of Poseidon: The model was pretrained with periodic boundary conditions and thus initially considers Mars as a torus; to match the training data we interpolated our data to resolution $128 \times 128$; there is no knowledge of forcings or boundary conditions and no way to infer position- or time-dependent characterstics as standard in weather models. We did a rather crude mapping when assigning temperature to the density channel. Finally, there is of course the fact that the pretraining data had no knowledge of the third dimension.

Now, to avoid misunderstanding, the above is not meant to critique the architecture choices of Poseidon. But rather to illustrate differences between PDE-FMs and models for atmospheric physics. One should be able to obtain considerably more performant neural emulators for weather when the underlying PDE-FM has been designed with this application in mind. The work of \citet{mccabe2025walrus} for example is based on the Well dataset \citep{ohana2024well} which includes both three and four dimensional data. It also includes parameters such as temperature. Other questions would be the inclusion of forcings and accounting for curved (i.e.~spherical) geometry.

\paragraph{Future work} Our results show that PDE foundation models can reduce data and compute requirements and improve accuracy in emulators of Mars atmospheric dynamics in data-limited settings. This suggests that PDE foundation model may also help in other data-limited settings such as regional weather on Earth. Crucially, our results point towards avenues to improve PDE-FMs for this and other real-world use cases. Given the considerations above, graph models with a global/local multi-scale structure might be of particular interest.

\bibliography{mars_pde}
\bibliographystyle{abbrvnat}

\appendix
\section{Additional results}

\begin{figure}[htbp]
\begin{center}
\includegraphics[width=0.5\textwidth]{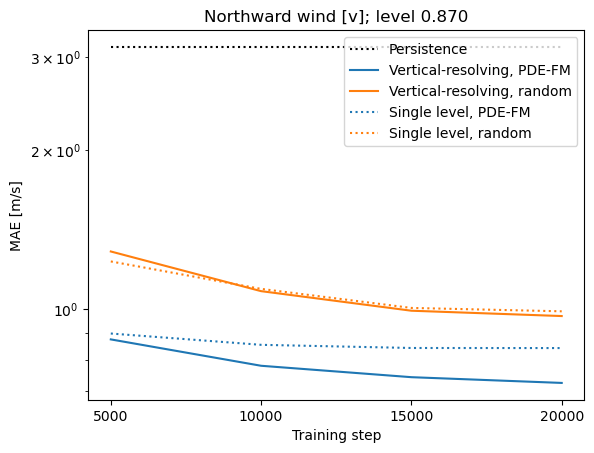}  
\end{center}
\caption{Impact of pretraining and dimensional extension on physical quantities; northward wind.}
\label{fig:mars_3d_forecast_impact_northward_wind}
\end{figure}

Figure \ref{fig:mars_3d_forecast_impact_northward_wind} shows the MAE for the v component of wind for the single level and vertical-resolving models. It is the complement of figure \ref{fig:mars_3d_forecast_impact_eastward_wind}.

\end{document}